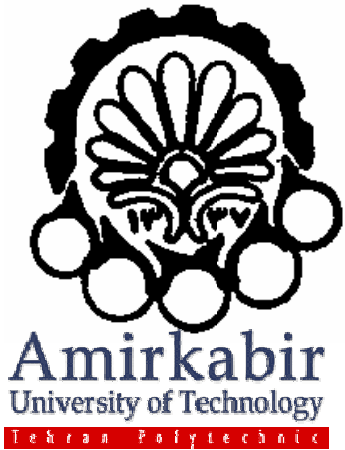

*A Brief Report on:*

# Development of Hybrid Intelligent Systems and their Applications from Engineering Systems to Complex Systems

**Hamed Owladeghaffari**

*Department of Mining &Metallurgical Engineering*

*Amirkabir University of Technology (Tehran Polytechnic)*

*Tehran, Iran*

*Sep 2006-May 2008*

**_(This version has been prepared for 10th Young Khwarizmi Award)_**

# Contents

**Abstract**

In this study, we introduce general frame of MAny Connected Intelligent Particles Systems (MACIPS). Connections and interconnections between particles get a complex behavior of such merely simple system (system in system).Contribution of natural computing, under information granulation theory, are the main topic of this spacious skeleton.

Upon this clue, we organize different algorithms involved a few prominent intelligent computing and approximate reasoning methods such as self organizing feature map (SOM)[9], Neuro-Fuzzy Inference System[10] ,Rough Set Theory (RST)[11], collaborative clustering ,Genetic Algorithm and Ant Colony System. Upon this, we have employed our algorithms on the several engineering systems, especially emerged systems in Civil and Mineral processing. In other process, we investigated how our algorithms can be taken as a linkage of government-society interaction, where government catches various fashions of behavior: "solid (absolute) or flexible". So, transition of such society, by changing of connectivity parameters (noise) from order to disorder is inferred. Add to this, one may find an indirect mapping among finical systems and eventual market fluctuations with MACIPS.

*In the following sections, we will mention the main topics of the suggested proposal, briefly <u>Details of the proposed algorithms can be found in the references</u>.*

# 1. Introduction

Complex systems are often coincided with uncertainty and order-disorder transitions. Apart of uncertainty, fluctuations forces due to competition of between constructive particles of system drive the system towards order and disorder. There are numerous examples which their behaviors show such anomalies in their evolution, i.e., physical systems, biological, financial systems [1]. In other view, in monitoring of most complex systems, there are some generic challenges for example sparse essence, conflicts in different levels, inaccuracy and limitation of measurements

,which in beyond of inherent feature of such interacted systems are real obstacle in their analysis and predicating of behaviors. There are many methods to analyzing of systems include many particles that are acting on each other, for example statistical methods [2], Vicsek model [3].

Other solution is finding out of "main nominations of each distinct behavior which may has overlapping, in part, to others". This advance is to bate of some mentioned difficulties that can be concluded in the "information granules" proposed by Zadeh [4]. In fact, more complex systems in their natural shape can be described in the sense of networks, which are made of connections among the units. These units are several facets of information granules as well as clusters, groups, communities, modules [5]. Let us consider a more real feature: dynamic natural particles in their inherent properties have (had have-will have) several appearances of "natural" attributes as in individually or in group forms. On the other hand, in society, interacting of main such characteristics (or may extra- natural forces: metaphysic) in facing of predictable or unpredictable events, determines destination of the supposed society.

Based upon the above, hierarchical nature of complex systems [6], developed (developing) several branches of natural computing (and related limbs) [7], collaborations [13], conflicts [11], emotions and other features of real complex systems, we propose a general framework of the known computing methods in the connected (or complex hybrid) shape, so that the aim is to inferring of the substantial behaviors of intricate and entangled large societies. Obviously, connections between units of computing cores (intelligent particles) can introduce part (or may full) of the comportments (demeanors-deportments...).

Complexity of this system, called MAny Connected Intelligent Particles Systems (MACIPS), add to reactions of particles against information flow, and can open new horizons in studying of this big query: is there a unified theory for the ways in which elements of a system(or aggregation of systems) organize themselves to produce a behavior?[8]. With expanding of a few MACIPS (Fig.1.) within a network (or complex network), we may construe events of our world within small world. Considering of growing, evolution, cliquing, competition and collaboration among supposed networks can instill a concomitant strategy on the insatiable problems of our world (Fig.2.).

In this study, we select a few limited parts of MACIPS, as well as hybrid intelligent systems, and investigate several levels of responses in facing of real information. We show how relatively such our simple methods that can produce (mimic) complicated behavior such government- nation interactions. Mutual relations between algorithms layers identify order-disorder transferring of such systems. So, we found our proposed methods have good ability in prediction and controlling of some engineering systems as well as those are emerged in Dam engineering, Geoseicince, Mineral processing. So, based on the mentioned system, we have developed a general intelligent rock engineering called: INtelligent Rock Engineering System (INRES). Developing of such intelligent hierarchical networks, investigations of their performances on the noisy information and exploration of possible relate between phase transition steps of the MACIPS and flow of in formations are new interesting fields, as well in various fields of science and economy.

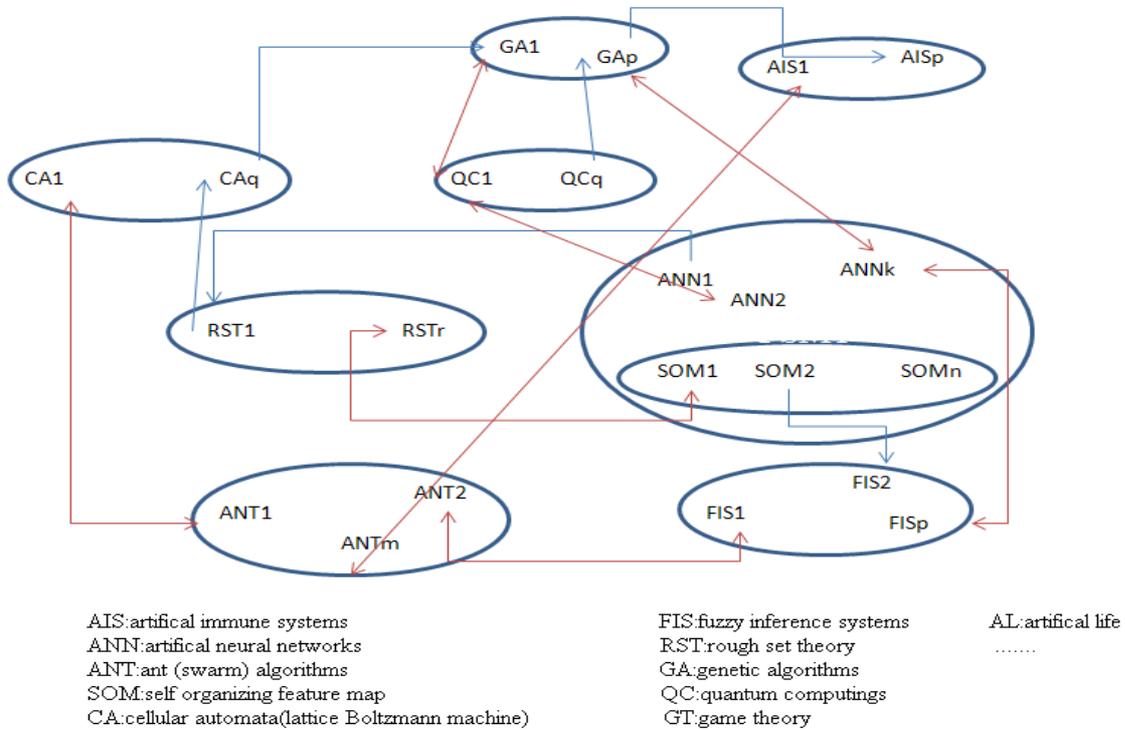

**Figure 1. A schematic view of MAny Connected Intelligent Particles Systems (MACIPS)**

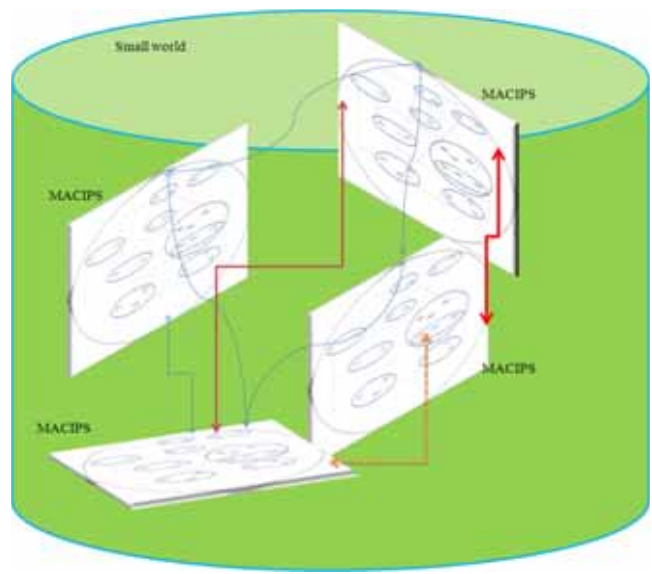

**Figure 2. A schematic Small world perspective: interactions of some MACIPS within network**

## 2. The proposed Algorithms

**2-1- A general methodology in designing:** *INtelligent Rock Engineering System (INRES)*:
See Details of the algorithms in the References.

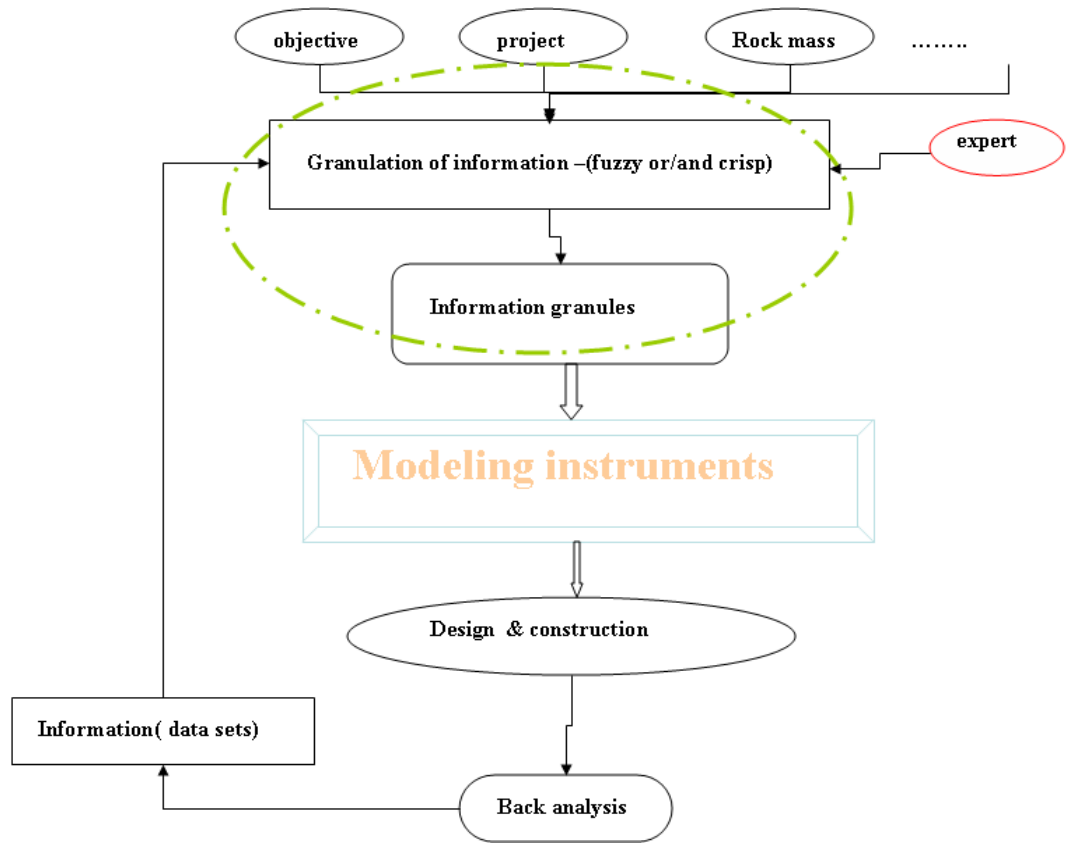

**Figure 3. A general methodology for Rock Engineering Design using Information granulation theory &extended Modeling instruments ([16], [17], [18], [21] & [22])**

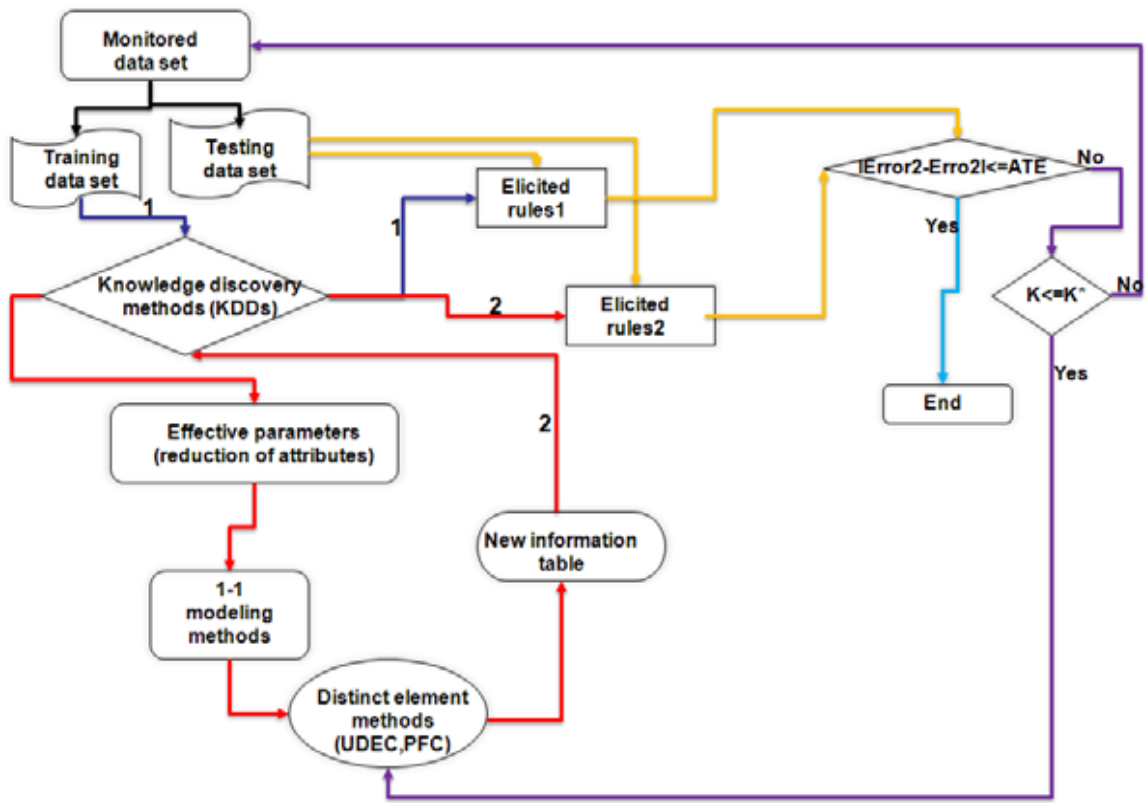

**Figure 4. Bridging between Knowledge Discovery Methods & Distinct Element Methods)-[23], [24], [26], [27] & [28]**

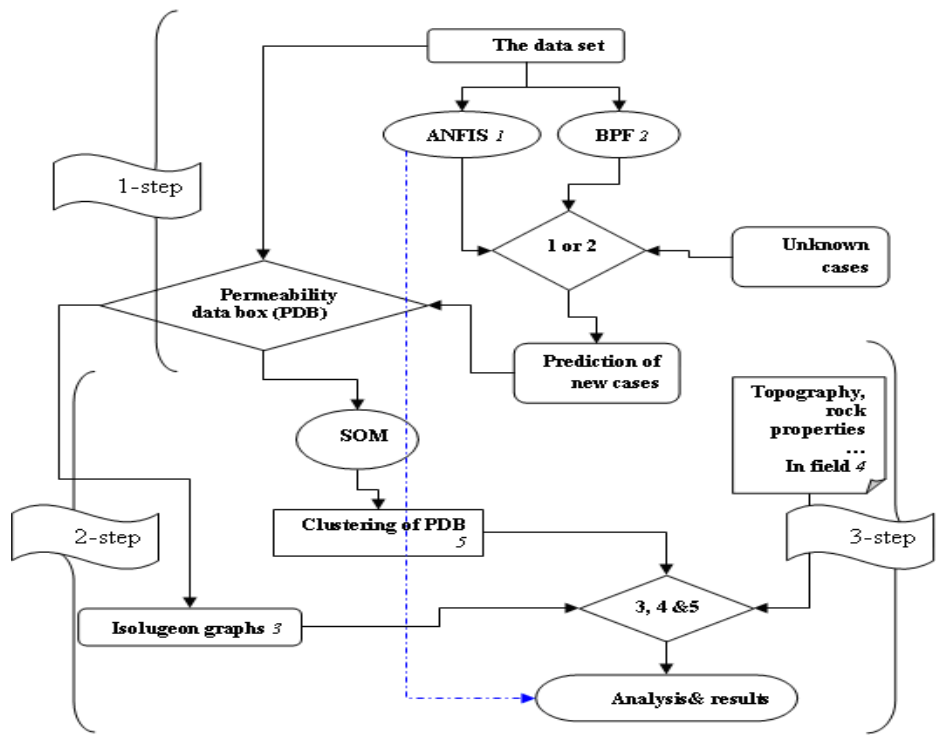

**Figure 5. A strategy in "Modeling Instrument" to Permeability Analysis "-[25]**

## 2-2-Developed Algorithms:

Developed algorithms in Fig (6-10) use four basic axioms upon the balancing of the successive granules assumption:

- Step (1): dividing the monitored data into groups of training and testing data
- Step (2): first granulation (crisp) by SOM or other crisp granulation methods
  Step (2-1): selecting the level of granularity randomly or depend on the obtained error from the NFIS or RST (regular neuron growth)
  Step (2-2): construction of the granules (crisp).
- Step (3): second granulation (fuzzy or rough granules) by NFIS or RST
  Step (3-1): crisp granules as a new data.
  Step (3-2): selecting the level of granularity; (Error level, number of rules, strength threshold...)
  Step (3-3): checking the suitability. (Close-open iteration: referring to the real data and reinspect closed world)
  Step (3-4): construction of fuzzy/rough granules.
- Step (4): extraction of knowledge rules

Selection of initial crisp granules can be supposed as "Close World Assumption (CWA)" .But in many applications, the assumption of complete information is not feasible, and only cannot be used. In such cases, an "Open World Assumption (OWA)', where information not known by an agent is assumed to be unknown, is often accepted [13].

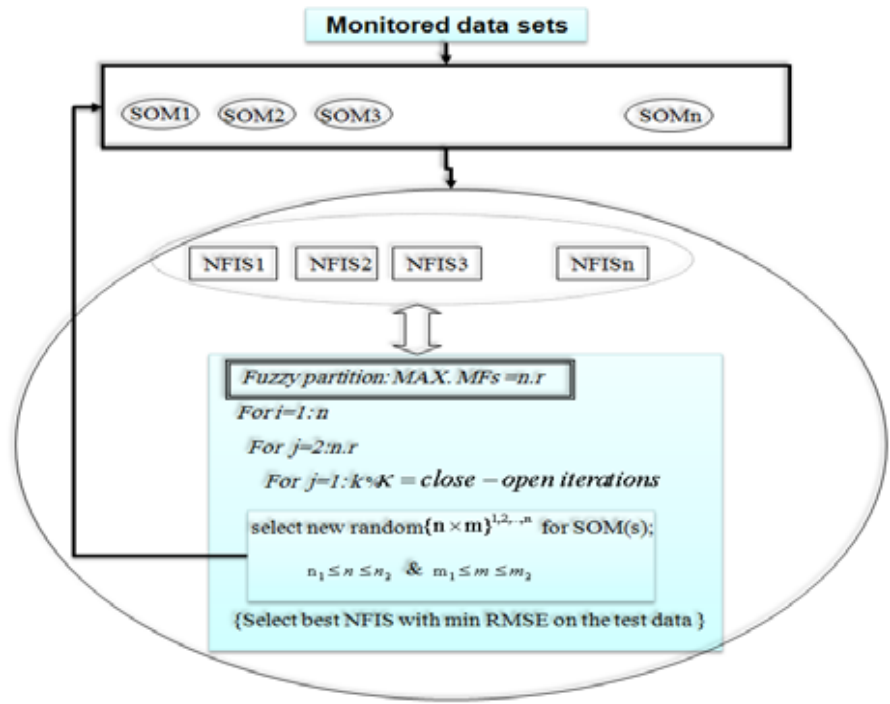

**Figure 6. Self Organizing Neuro-Fuzzy Inference System (SONFIS) – ([14] - [22])**

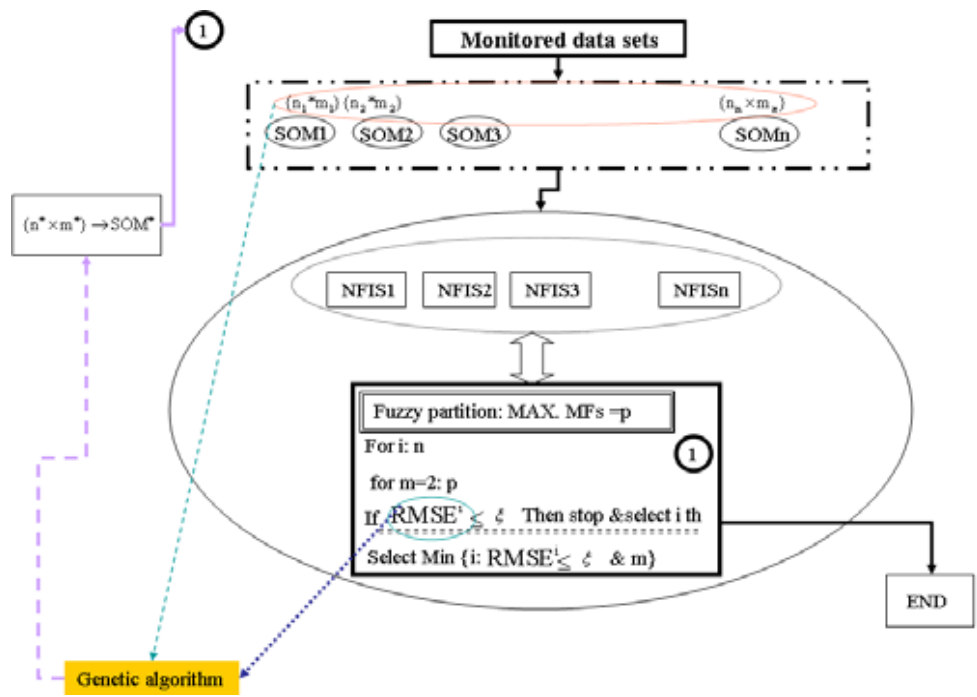

Figure 7. Evolutionary Self Organizing Neuro-Fuzzy Inference System (E-SONFIS)-[29]

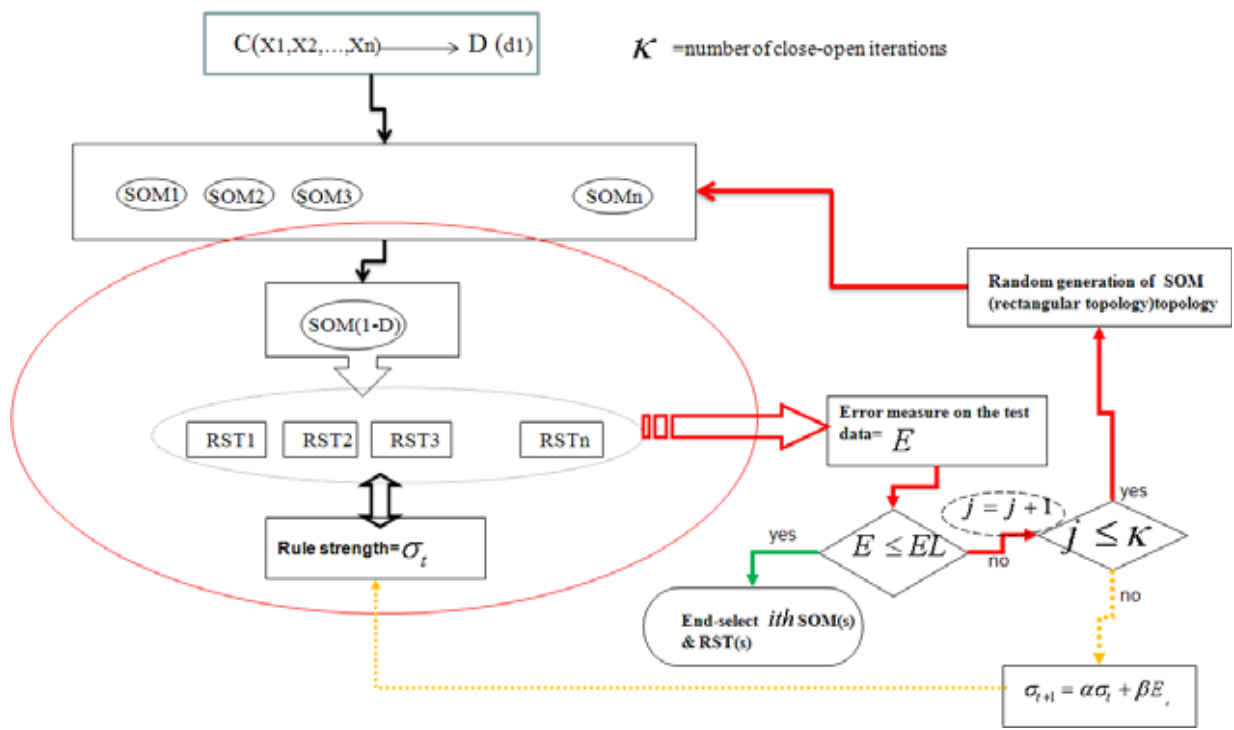

Figure 8. Self Organizing Rough Set Theory-Random neuron growth & adaptive strength factor (SORST-R)-[15], [17], [19],[20]

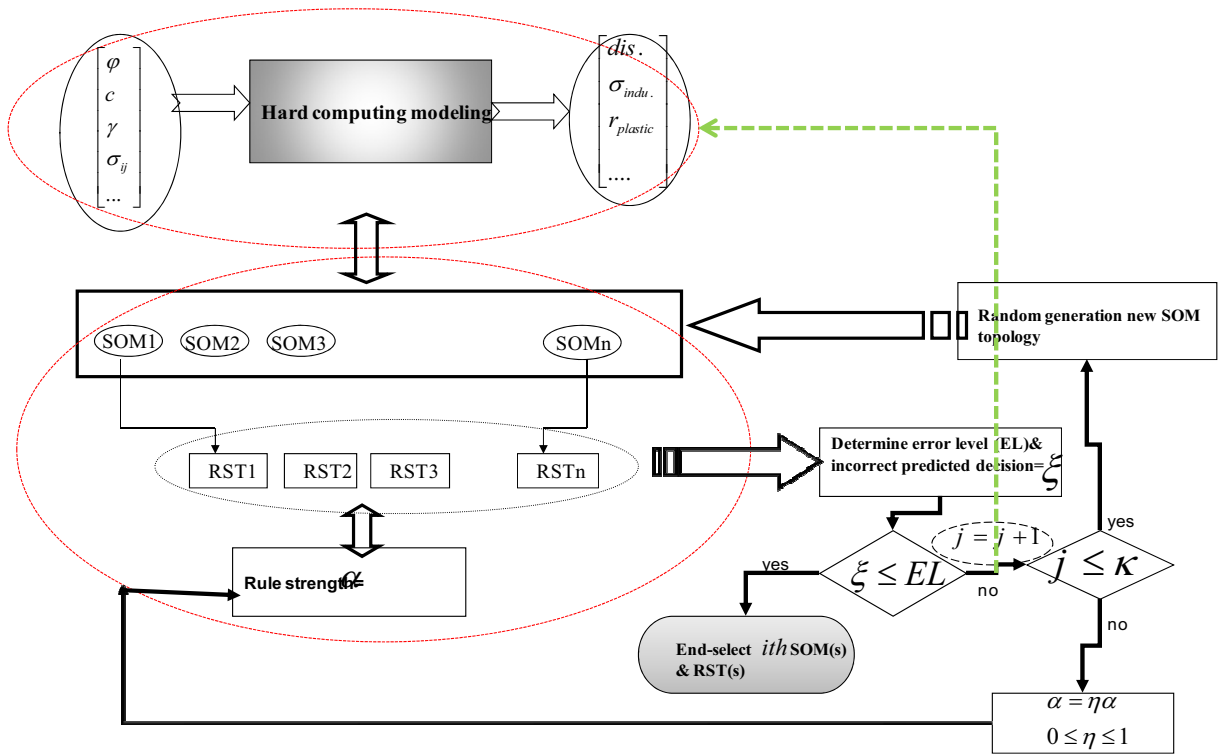

**Figure 9 . Bridging of hard computations and Self Organizing Rough Set Theory-Random neuron growth & adaptive strength factor (SORST-R)-[17], [22]**

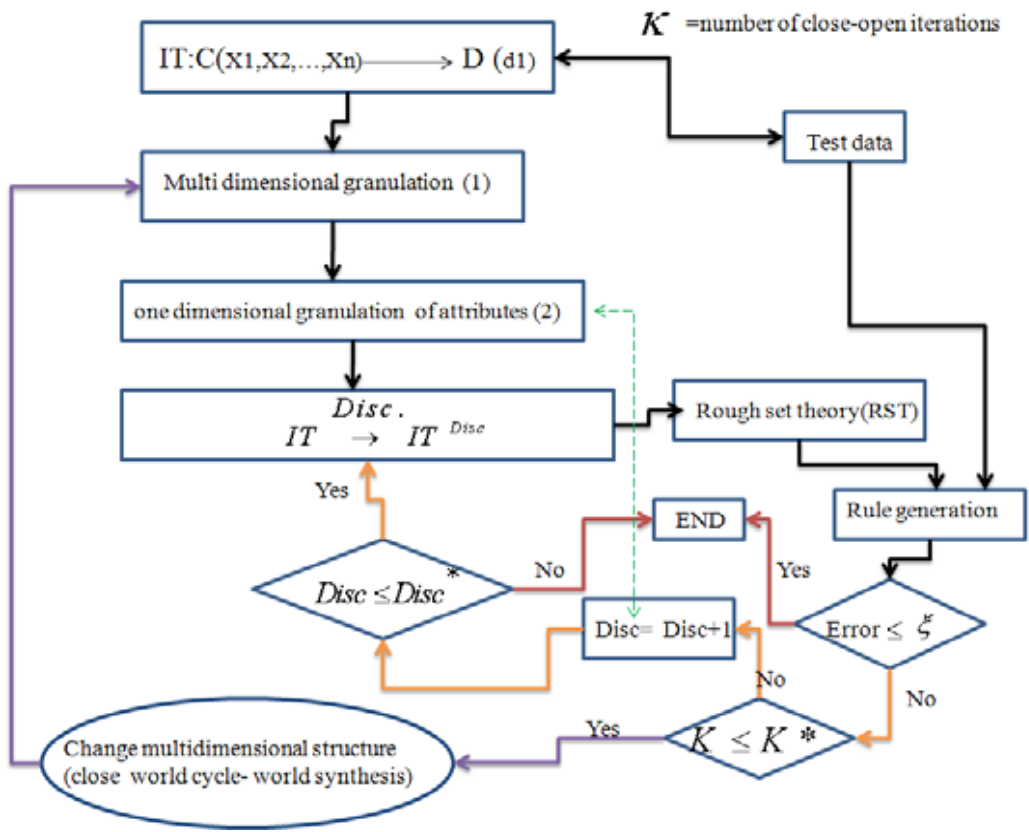

**Figure 10 .Self Organizing Rough Set Theory-Adaptive Scaling (SORST-AS)-[14]**

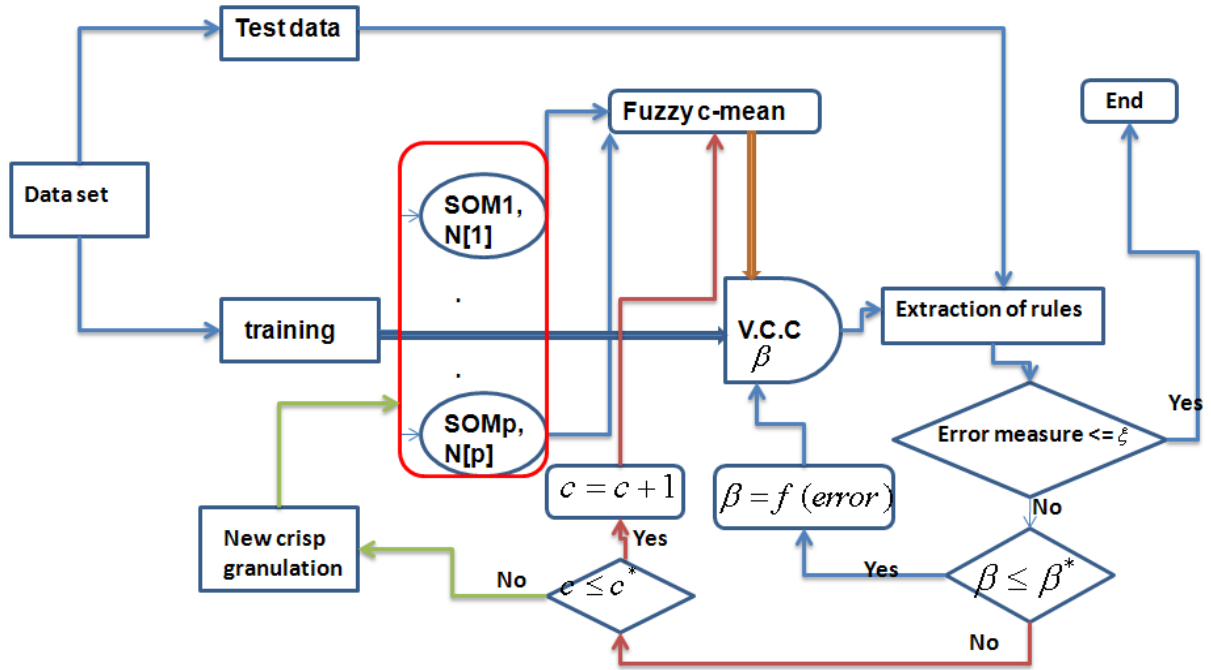

**Figure 11. a combining between vertical collaborative clustering (VCC) [13] &balancing of granules using ant colony system in the granular worlds [30]**

Details of Fig11:

Step1- dividing the monitored data in to groups: training and testing sets.

Step2-initial granulation (crisp) by $SOM^1,\ldots,SOM^p$ and $N[1],\ldots N[ii]; i=1,\ldots,p$; With different 2-Dstructure (Randomly-by training data set).

Step3-fuzzy C-mean algorithm (or other): creation **C** clusters (number of clusters) for each $N[ii]$.

Step 4-VCC algorithm

Step 4-1- set $\beta[i,j]$, randomly.

Step4-2-finding prototypes for *i=1,…p* and extraction of rules( linear format).

Step4-3- determine $\Delta\varepsilon_i$ (by testing data).

Step4-4- $|\Delta\varepsilon_i| \leq \xi$, if yes------→end, else go to 5

Step5-Ant colony system, set $t^*$ (repeat $t \leq t^*$)

5-1- set $\overline{\Delta \varepsilon}^t_i = \dfrac{\Delta \varepsilon^t_i}{\sum_{i=1}^{p} \Delta \varepsilon^t_i}$

5-2- $\dfrac{1}{\left|\overline{\Delta \varepsilon}_i^t - \overline{\Delta \varepsilon}_j^t\right|} = \Delta \tau_{ij}^t$

5-3- $\tau_{ij}(t+1) = (1-\rho)\tau_{ij}^t + \Delta \tau_{ij}^t$

$$\beta = \lambda \tau_{ij}$$

==================================================

Else (if $t \geq t^*$)

(Start of 1-close-open iteration or 2-open-close iteration)

1- set $k \leq k^*$ & $\max c = c^*$ (repeat)

Change randomly, crisp granules ( 1,…p),in SOM core.

Go to step 2

Else *c=c+1*

*If $c \leq c^*$* ,go to step 3.

2- Else (if $t \geq t^*$ )& *If $c \leq c^*$*,

Set c=c+1 and go to step 3.

Else, Change randomly, crisp granules ( 1,…p),in SOM core.

Go to step 2

-------------------------------------------------

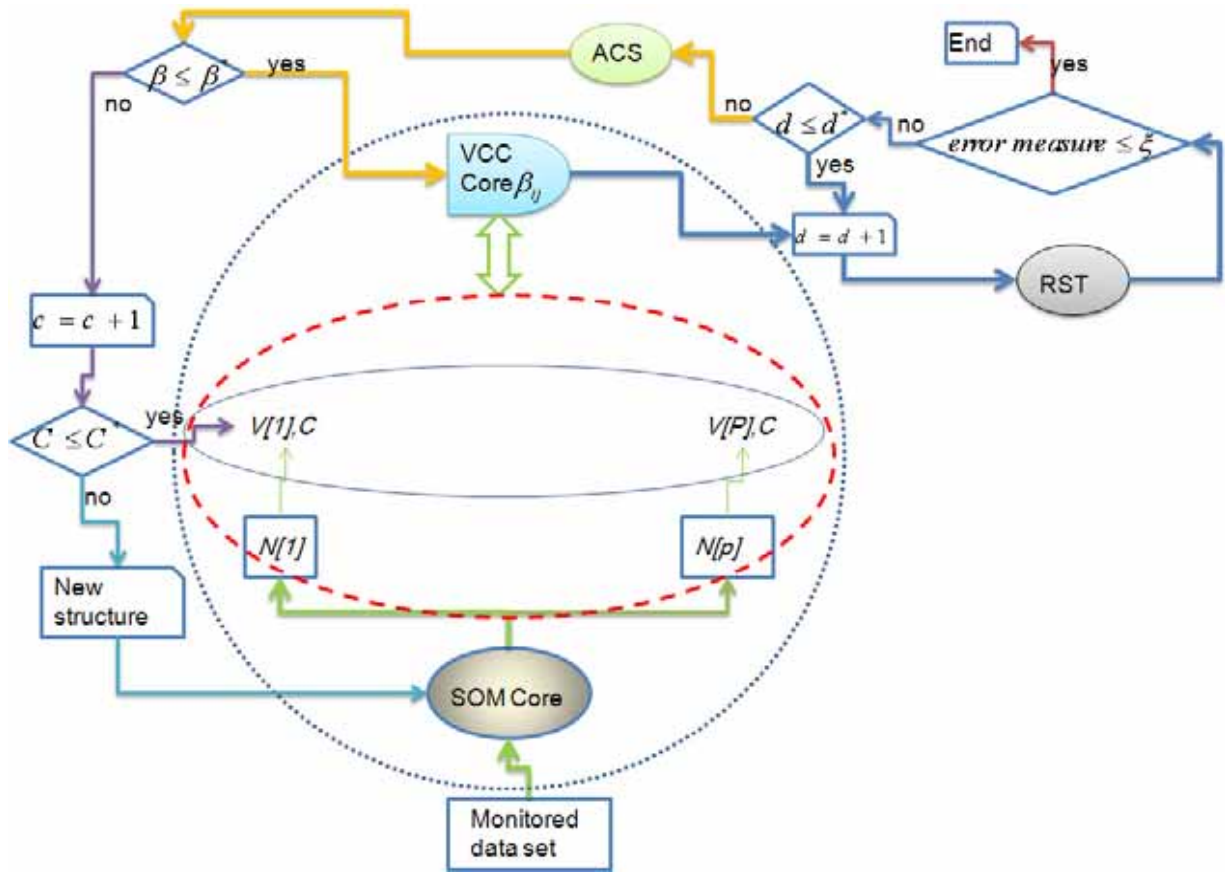

**Figure 12. A self organizing Vertical Collaborative Clustering (VCC) upon the balancing of granules, Rough Set Theory and ant colony system in the granular worlds [30]**

7. Castro,L.N.D.: Fundamentals of Natural Computing: An Overview. Physics of Life Reviews.4, pp1-36(2006)
8. Vicsek, T.: The Bigger Picture. Nature.418, 131(2002)
9. Kohonen, T.: Self-Organization and Associate Memory. 2nd edit. Springer – Verlag, Berlin (1983)
10. Jang, J., S. R., Sun, C. T., Mizutani, E.: Neuro-Fuzzy and Soft Computing", Newjersy, Prentice Hall (1997)
11. Pawlak, Z.: Rough Sets: Theoretical Aspects Reasoning about Data. Kluwer academic, Boston (1991)
12. Doherty, P., Kachniarz, J., Szatas, A.: Using Contextually Closed Queries for Local Closed-World Reasoning in Rough Knowledge Data Base. In rough-neural computing techniques for comparing with words, eds. Pal, S. K., Polkowski, L., Skowron, A. pp.219—250(2004).
13. Pedrycz, W.: Collaborative fuzzy clustering. Pattern Recognition Letters **23**;1675–1686, (2002)

## Publications on the Proposal:

(Marked Documents are Available at: http://arxiv.org/find/all/1/all:+Owladeghaffari/0/1/0/all/0/1 -So see the annexed CD-ROM)

**A) Journals**

- An Intelligent Algorithm to Permeability Analysis, H.Owladeghaffari, E.Bakhtavar& K.Shahriar, Bulletin of Engineering Geology and the Environment, in revising

- An Application of Soft Granulation Theory to Permeability Analysis , H. Owladeghaffari, M.Sharifzadeh ,K.Shahriar & W. Pedrycz, Int. Journal of Rock Mechanics and Mining Sciences, In revising

- Modeling of Social phase Transitions Using Intelligent Systems, H.Owladeghaffari & K.Shahriar submitted to Journal of Systems Science and Complexity, May 2008

- Interactions of Nations-Governments Using Intelligent Systems; H.Owladeghaffari ;K.Shahriar and W. Pedrycz ; submitted to Applied Soft Computing, May 2008

**B) Conference Papers &Preprints**

14. *Order to Disorder Transitions in Hybrid Intelligent Systems: a Hatch to the Interactions of Nations -Governments; H.Owladeghaffari; The 2008 IEEE International Conference on Granular Computing (GrC 2008), China, Aug 2008
15. *Phase Transition in SONFIS&SORST; H.Owladeghaffari; W. Pedrycz; The Sixth International Conference on Rough Sets and Current Trends in Computing; Akron , Ohio, USA,2008

16. *Graphical Estimation of Permeability Using RST&NFIS ;H.Owladeghaffari ; K.Shahriar &W. Pedrycz; The 27th Annual Meeting of the North American Fuzzy Information Processing Society (NAFIPS'08) - New York, NY, USA, May 19-22, 2008

17. *Rock Mechanics Modeling Based on Soft Granulation Theory; H.Owladeghaffari, M.Sharifzadeh ,K.Shahriar & E.Bakhtavar; 42nd U.S. Rock Mechanics Symposium 2nd U.S.-Canada Rock Mechanics Symposium, 2008(in press)

18. *Permeability Analysis Based on Information Granulation Theory; M.Sharifzadeh, H.Owladeghaffari , E.Bakhtavar ,K.Shahriar; The 12th International Conference of International Association for Computer Methods and Advances in Geomechanics (IACMAG)1-6 ;October, 2008

19. *Analysis of Hydrocyclone Performance Based on Information Granulation Theory; Hamed O.Ghaffari, Majid Ejtemaei & Mehdi Irannajad; 8th. World Congress on Computational Mechanics (WCCM8); 5th. European Congress on Computational Methods in Applied Sciences and Engineering (ECCOMAS 2008); June 30 – July 5, 2008-Venice, Italy

20. Knowledge Discovery of Hydrocyclone's Circuit Based on SONFIS&SORST; Hamed O.Ghaffari , Majid Ejtemaei & Mehdi Irannajad;11th International Mineral Processing Symposium-21-23 October-Antalya- Turkey

21. *Assessment of Effective Parameters on Dilution Using Approximate Reasoning Methods in Longwall Mining Method ; Iran Coal Mines; H.O.Ghaffari, K.Shahriar & Gh.Saeedi; The 21st World Mining Congress &EXPO 2008; 7-11 ;September 2008,;Krakow, Poland

22. Back Analysis Based on SOM-RST System; H.Owladeghaffari; H.Aghababaei ; Proceedings of the 10th Internatinoal Symposium on Landslides and Engineered Slopes, 30 June – 4 July 2008, Xi'an, China ;Editor(s) - Zu-yu Chen, Jian-Min Zhang, Ken Ho, Fa-Quan Wu, Zhogn-Kui Li

23. *Toward Fuzzy Block Theory ; h.owladeghaffari, h.salari-rad ;PROCEEDINGS OF THE 5TH INTERNATIONAL WORKSHOP ON APPLICATIONS OF COMPUTATIONAL MECHANICS IN GEOTECHNICAL ENGINEERING GUIMARÃES / PORTUGAL / 1-4 APRIL 2007-CDROM

24. *Contact State Analysis Using NFIS &SOM; h.owladeghaffari; COMPUTATIONAL MECHANICS; ISCM2007, July 30- August 1, 2007, Beijing,China ;2007- CDROM.

25. Analysis of Permeability Using BPF, ANFIS & SOM; K. Shahriar& H. Owladeghaffari; 1st Canada-U.S. Rock Mechanics Symposium -Rock Mechanics: Meeting Society's Challenges and Demands, Vancouver, Canada, 27-31 May 2007 ; Editor(s) : Erik Eberhardt ,Doug Stead &Tom Morrison pp.303-7, 2007

**C) Non-refereed Papers**

26. Analysis of Key Blocks Stability Using FIS&SOM; H. Owladeghaffari; fall 2006
27. Parallelization of DDA Using Soft Computing Approaches; H. Owladeghaffari; winter 2006
28. towards Fuzzy Analysis on Contacts in Block System; H. Owladeghaffari; winter 2007
29. SONFIS &SORST based on Genetic algorithm ;Winter 2008
30. A self organizing Vertical Collaborative Clustering; W. Pedrycz & H.Owladeghaffari

---------------------------------------------------------------------------------------------------------------------

**Hamed Owladeghaffari**: *Department of Mining and Metallurgical Engineering; Amirkabir University of Technology (Tehran Polytechnic)-Graduate education center*-Email: h.o.ghaffari@gmail.com